\documentclass[balance,upint,subscriptcorrection,varvw,mathalfa=cal=boondoxo,spanish,french,vietnamese,russian,greek,pdf-a,fontspec,colorlinks]{asmeconf}

\usepackage{packages}
\usepackage{comment}
\usepackage{afterpage}

\hypersetup{%
	pdfauthor={Daniel Menges},
	pdftitle={Digital Twin of Autonomous Surface Vessels for Safe Maritime Navigation Enabled through Predictive Modeling and Reinforcement Learning⁠},                  
	pdfkeywords={Digital Twin, Autonomous Surface Vessel, Situational Awareness, Target Tracking, Predictive Safety Filter, Reinforcement Learning},
}

\begin{document}
	
	

    \title{Digital Twin of Autonomous Surface Vessels for Safe Maritime Navigation Enabled through Predictive Modeling and Reinforcement Learning} 
	%
	%
	%
	\SetAuthors{%
		Daniel Menges\affil{1}\CorrespondingAuthor{},
            Andreas Von Brandis\affil{1},
		Adil Rasheed\affil{1}\affil{2}
	}	
	\SetAffiliation{1}{Norwegian University of Science and Technology, Trondheim, Norway}
        \SetAffiliation{2}{Department of Mathematics and Cybernetics, SINTEF Digital, Trondheim, Norway}
	\maketitle
	\begingroup
	\renewcommand
	\thefootnote{\textsection}
	\endgroup	
	
	\normalfont
	

	\begin{abstract}
        Autonomous surface vessels (ASVs) play an increasingly important role in the safety and sustainability of open sea operations. Since most maritime accidents are related to human failure, intelligent algorithms for autonomous collision avoidance and path following can drastically reduce the risk in the maritime sector. A DT is a virtual representative of a real physical system and can enhance the situational awareness (SITAW) of such an ASV to generate optimal decisions. This work builds on an existing DT framework for ASVs and demonstrates foundations for enabling predictive, prescriptive, and autonomous capabilities. In this context, sophisticated target tracking approaches are crucial for estimating and predicting the position and motion of other dynamic objects. The applied tracking method is enabled by real-time automatic identification system (AIS) data and synthetic light detection and ranging (LiDAR) measurements. To guarantee safety during autonomous operations, we applied a predictive safety filter to correct inputs from a reinforcement learning-based controller. The approaches are implemented into a DT built with the Unity game engine. As a result, this work demonstrates the potential of a DT capable of making predictions, playing through various what-if scenarios, and providing optimal control decisions according to its enhanced SITAW.
	\end{abstract}
	\begin{nomenclature}[5em]
		\EntryHeading{Abbreviations}
            \entry{3D}{Three-dimensional}
            \entry{AIS}{Automatic identification system}
            \entry{API}{Application programming interface}
            \entry{ASV}{Autonomous surface vessel}
            \entry{CAD}{Computer-aided design}
            \entry{COLAV}{Collision avoidance}
            \entry{CTE}{Cross-track error}
            \entry{DNN}{Deep neural network}
            \entry{DRL}{Deep reinforcement learning}
		\entry{DT}{Digital twin}
            \entry{EKF}{Extended Kalman filter}
            \entry{GNSS}{Global navigation satellite system}
            \entry{IMU}{Inertial measurement unit}
            \entry{IoT}{Internet of things}
            \entry{KF}{Kalman filter}
            \entry{LiDAR}{Light detection and ranging}
            \entry{MLR}{Multiple linear regression}
            \entry{MPC}{Model predictive control}
            \entry{NMPC}{Nonlinear model predictive control}
            \entry{OCP}{Optimal control problem}
            \entry{PPO}{Proximal Policy Optimization}
            \entry{PSF}{Predictive safety filter}
            \entry{PTT}{Predictive target tracking}
            \entry{RADAR}{Radio detection and ranging}
            \entry{RL}{Reinforcement learning}
            \entry{SDP}{Semidefinite program}
            \entry{SITAW}{Situational awareness}
            \entry{UKF}{Unscented Kalman filter}
	\end{nomenclature}
	
	
\section{Introduction}
Given that about 90\% of the world's traded goods are carried by cargo ships \cite{Wang2020aso}, the introduction of autonomous surface vessels (ASVs) entails the potential to improve safety in the maritime environment by parallel reducing CO$_2$ emissions to counteract against the global warming. However, the successful implementation of ASVs inheres to many challenges, including navigation control, multivariate data analysis, communication, and Internet of Things (IoT) resource orchestration \cite{Martelli2021aoo}. A key factor in such autonomous systems is their situational awareness (SITAW) \cite{Thombre2022} for assessing the risk in connection with intelligent control algorithms for optimized path following and collision avoidance \cite{Vagale2021ppa}. Given these interconnected and multidimensional challenges, the utilization of modern techniques is inevitable.

Digital twins (DTs) offer new opportunities for our growing digitalization and its various applications in different fields. In general, a DT represents a digital version of a real physical system or process enabled through data to allow capabilities such as real-time prediction, optimization, monitoring, control, and improved decision making \cite{Rasheed2020dtv}. Since the definition and development of DTs is still in an initial phase, we introduce a capability scale described in \cite{San2021haaa}, classifying its proficiency into 0-standalone, 1-descriptive, 2-diagnostic, 3-predictive, 4-prescriptive, and 5-autonomous. The following section introduces the idea behind this concept in more detail. 
As stated in the review on DTs in the maritime domain \cite{Madusanka2023dti}, the potential of DTs will revolutionize the maritime sector, especially in the field of shipping.

The number of proposed DT concepts related to ASVs increases. For instance, \cite{Heo2023aps} introduces a preliminary study for the development of a DT technology using an ASV. However, while data acquisition is performed for subsequent analysis and modeling, the proposal did not implement a real-time connection from the physical system to its DT. In addition, in \cite{Riordan2023sag}, an anti-grounding scheme is demonstrated for ASVs assisted by a DT using an electronic chart display and information system. The proposal of \cite{Vasanthan2021csl} presents the combination of supervised learning and DTs concerning autonomous path planning. Furthermore, \cite{Hasan2023pdt} introduces a predictive DT for ASVs by utilizing a nonlinear adaptive observer and exogenous Kalman filters for state and parameter estimations while predicting future states using a discretized state space model. Predictive capabilities might be essential for real operations of a highly advanced DT of an ASV and its environment. However, predictions cover only a small subset of the potential that a DT can enable. Recently, a DT framework for an ASV was developed, described in \cite{Raza2022tid}, incorporating aspects such as SITAW using data from cameras, LiDAR, the global navigation satellite system (GNSS), and an inertial measurement unit (IMU). Furthermore, COLAV is approached by applying deep reinforcement learning (DRL), showing overall that a DT is a dynamic and safe concept for the design and development phase of ASVs. However, the proposal uses a highly simplified reward function to train the deep neural network (DNN), and the training phase considered only one static obstacle, making the proposed DRL control concept questionable in the case of more challenging scenarios.

In \cite{Menges2023dtf}, the development of a DT framework for an ASV is illustrated, considering the capability scale mentioned earlier. The study shows the development and concept of a versatile DT by setting the foundation of a three-dimensional (3D) digital environment with the Unity game engine using a real-world 3D elevation map and computer-aided design (CAD) models for the 3D representation of surface vessels. Furthermore, real-time automatic identification system (AIS) data and weather data are gathered to facilitate descriptive and diagnostic capabilities, while implemented nonlinear vessel dynamics define the kinetic behavior of the digital representatives. In addition, DRL is applied for autonomous path following and COLAV, where randomly generated scenarios, including highly dynamic obstacles, set the basis for training.
Summarized, it is shown how a DT of an ASV and its environment can generate situational awareness and enhance its perception space used for optimal control.

This work extends this existing framework through improved autonomous, prescriptive, and predictive capabilities. For this purpose, we introduce the theory and deployment of a predictive target tracking method enabling the estimation and prediction of the position and motion related to other dynamic objects using AIS data and synthetic light detection and ranging (LiDAR) measurements. In addition, we introduce the concept of a predictive safety filter (PSF) based on the theory of nonlinear model predictive control (NMPC) for safe control of ASVs. Both methodologies, novel with respect to DTs, are implemented in the extendable DT framework and finally depicted in the results section. Therefore, we introduce the necessary preliminaries in Section~\ref{sec:Theory}, followed by the novel methods and extensions of the DT described in Section~\ref{sec:Method}.

\section{Theory} \label{sec:Theory}
In this section, the concept of a DT in connection with the existing DT framework, introduced in \cite{Menges2023dtf}, is briefly explained. Moreover, the nonlinear vessel model and the concept of NMPC are described since they set the basis for the PSF deployed in the methodology section. Subsequently, a numerically stable ellipse fitting algorithm, the general concept of Kalman filters, and the idea of sensor fusion are introduced, setting the foundation for the predictive target tracking approach.

\subsection{Digital Twins} \label{sec:Digital_twins}
\begin{figure*}[htbp]
	\centering
    \includegraphics[width=\linewidth]{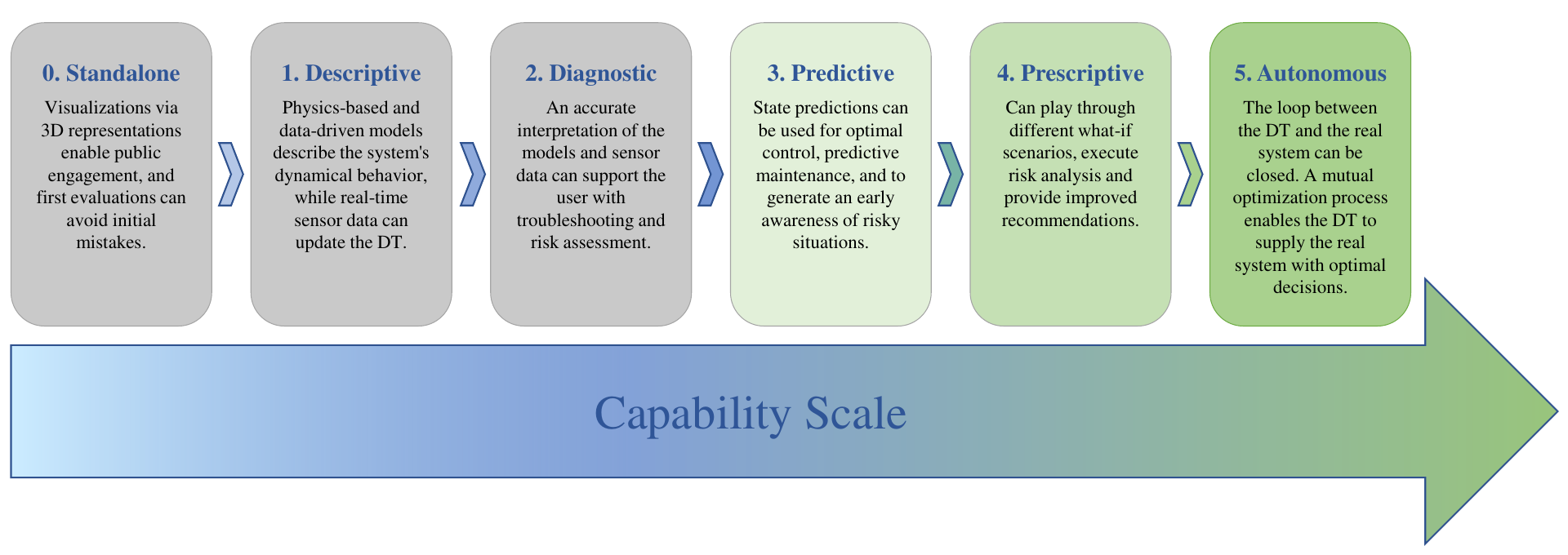}
	\caption{Capability scale of digital twins}
	\label{fig:DT_capability_levels}
\end{figure*}

As demonstrated in \cite{Menges2023dtf}, a digital twin (DT) can be characterized according to its set of capabilities. In general, a classification is defined by a scale from 0 to 5, as described below. Figure~\ref{fig:DT_capability_levels} shows the structure of the capability scale. Note that, depending on the application and its requirements, a DT can contain several combinations of capabilities given by 
\begin{equation}
    p = \sum_{k=1}^{n} \frac{n!}{k!(n-k)!}
\end{equation}
Since we classify into $n=6$ different categories, there are theoretically $p = 63$ various combinations. However, many capability levels build up on each other. Therefore, their general workflow is represented by an arrow and the capability levels covered in this work are highlighted in green.

\subsubsection*{Capability Level 0: Standalone}
\begin{figure}[htbp]
	\centering
    \includegraphics[width=\linewidth]{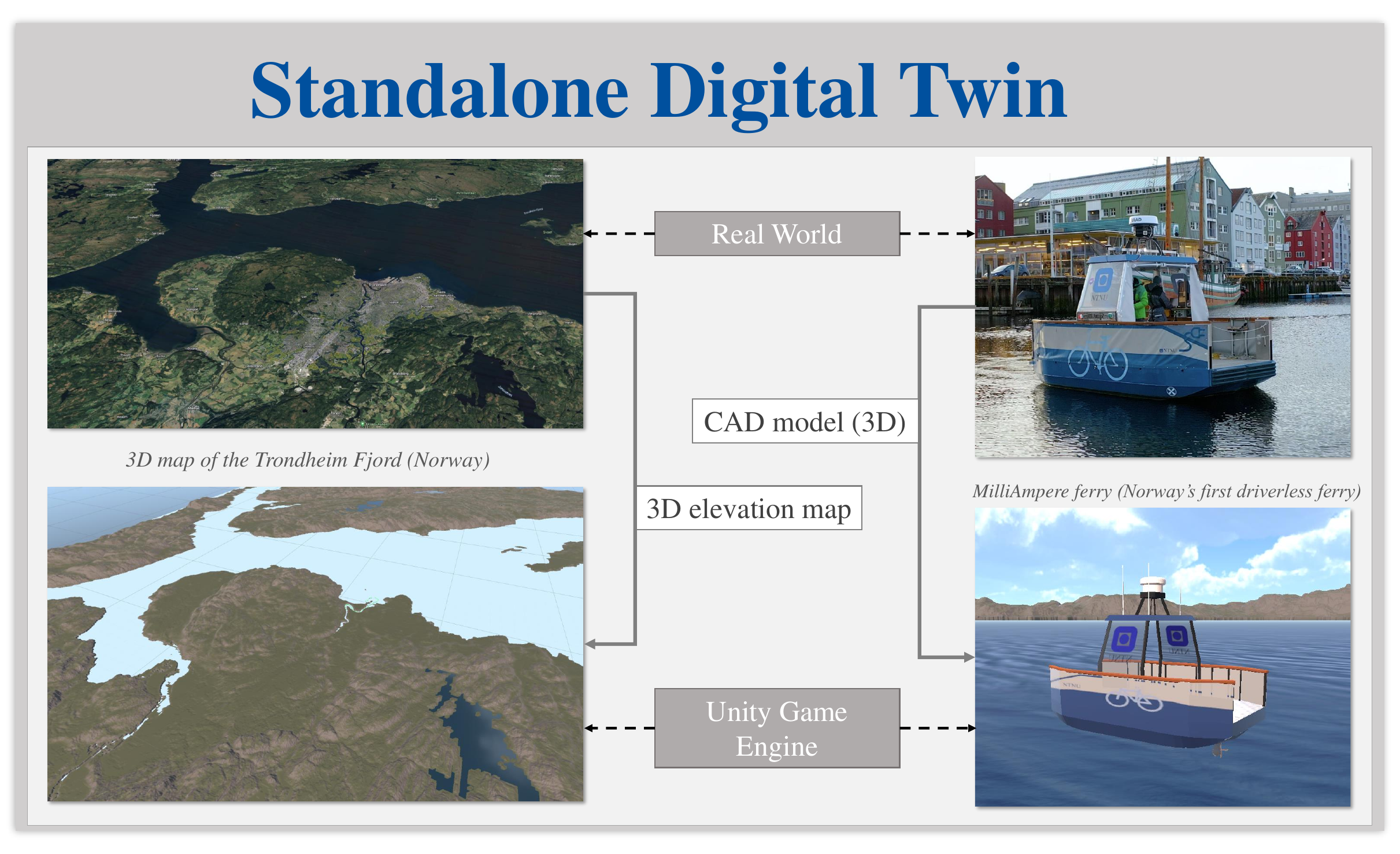}
	\caption{Implementation of a standalone digital twin in the Unity game engine.}
	\label{fig:Standalone_DT}
\end{figure}

A standalone DT serves as a visual demonstrator of its real representative. With regard to most physical systems, three-dimensional (3D) mappings are integrated into a DT. In the extendable DT framework, used in this work, CAD models of the vessels and a 3D elevation map of the Trondheim Fjord are implemented and represent the standalone model. In Fig.~\ref{fig:Standalone_DT}, the standalone DT implementation is depicted.

\subsubsection*{Capability Level 1: Descriptive}
\begin{figure}[htbp]
	\centering
	\includegraphics[width=\linewidth]{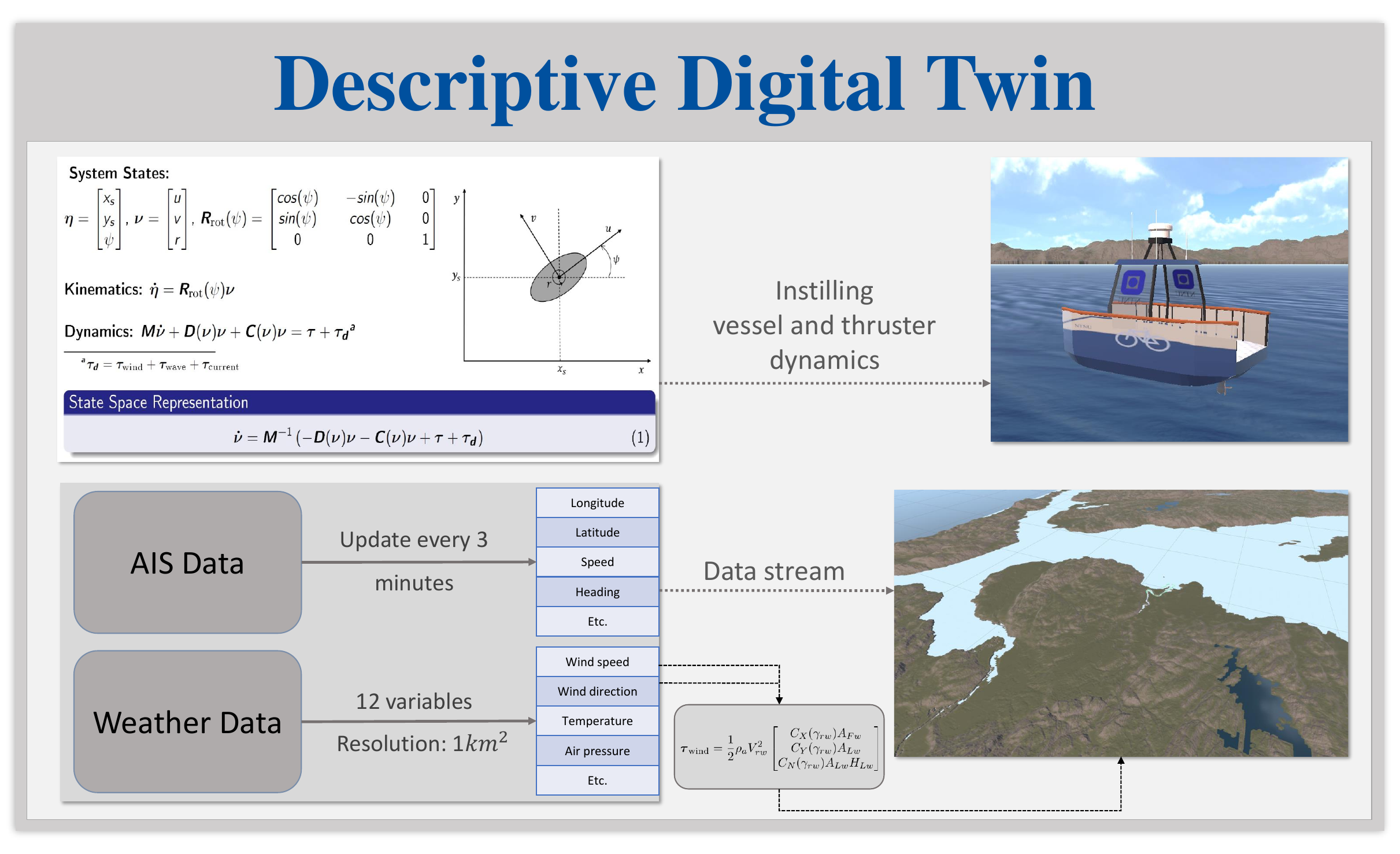}
	\caption{Implementation of a descriptive digital twin in the Unity game engine.}
	\label{fig:Descriptive_DT}
\end{figure}

The descriptive DT is based on physics-based and data-driven models. In parallel, it is driven by real-time data, enabling corrections of modeling errors. The existing framework uses nonlinear model and thruster dynamics of ASVs to describe their behavior. In addition, real-time AIS data is streamed, allowing mapping of the position and motion of other commercial ships. Since the AIS data are updated in time-varying intervals depending on message transmission, their motion within these intervals was extrapolated. Furthermore, weather data are streamed into the framework, containing information about wind speed, wind direction, temperature, air pressure, and more, with a resolution of one square kilometer. The implementation is shown in Fig.~\ref{fig:Descriptive_DT}

\subsubsection*{Capability Level 2: Diagnostic}
\begin{figure}[htbp!]
	\centering	\includegraphics[width=\linewidth]{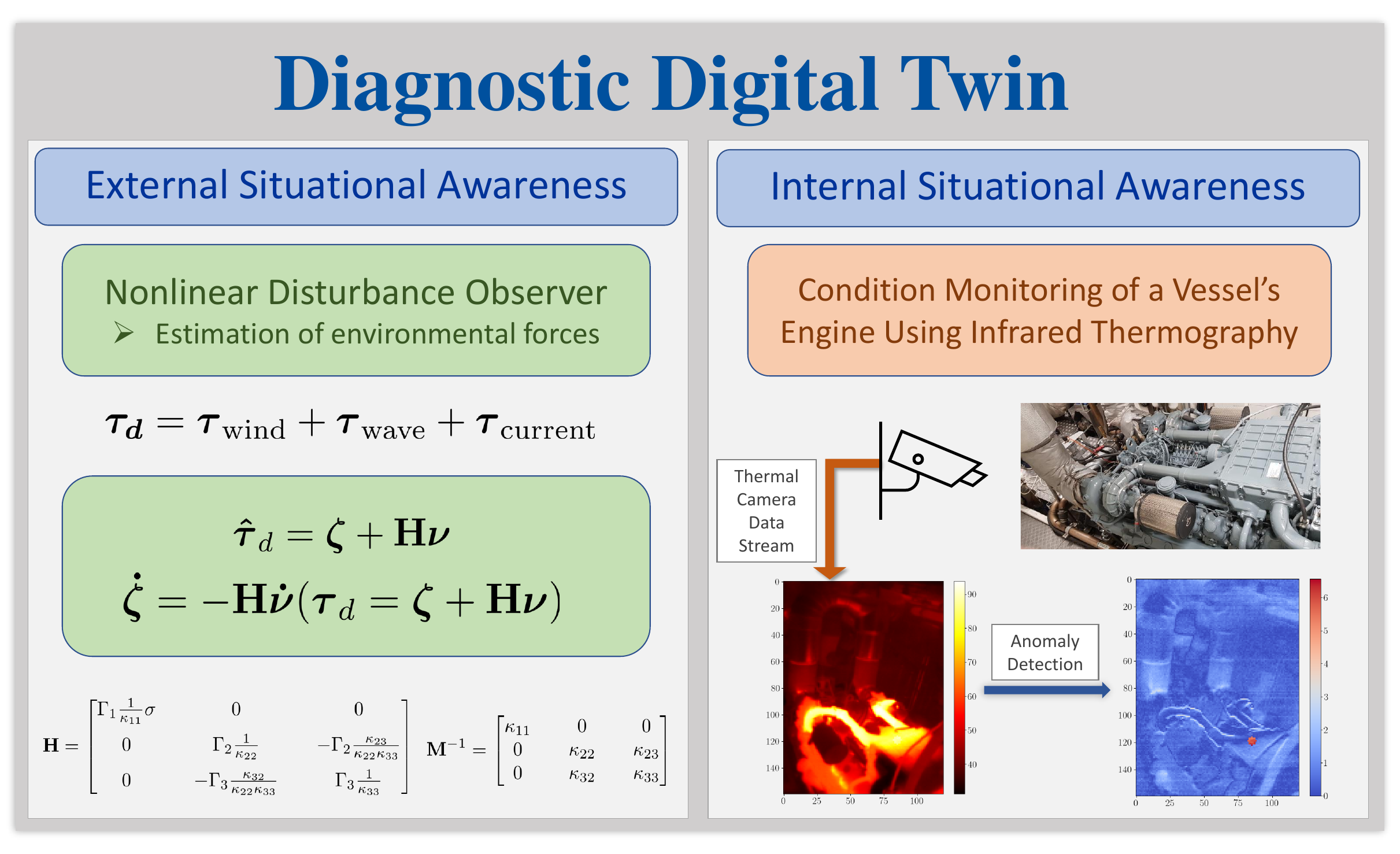}
	\caption{Implementation of a diagnostic digital twin in the Unity game engine.}
	\label{fig:Diagnostic_DT}
\end{figure}

Physics-based models and data streams enable the analysis of the situation and allow diagnostic statements.
The first diagnostic capabilities are instilled by a nonlinear disturbance observer for ASVs developed by \cite{Menges2023aed}. As a result, estimations of unknown environmental forces on an ASV impacted by wind, waves, and sea currents are accessible. These observed disturbances extend the SITAW of an ASV. This extended knowledge can subsequently be used for improved control considering the environmental impact. In addition, monitoring the condition of various components of the vessel is a crucial core competence considering capability level 2. This might be the detection of faults in an integrated power system, such as the diagnosis of propulsion branch faults \cite{Qi2022adt}. Initial solutions with regard to condition monitoring have already been developed. However, such internal SITAW of the ASV is not yet implemented in the DT framework, but the general idea of using thermal cameras to track the conditions of an engine is presented in Fig.~\ref{fig:Diagnostic_DT}. In summary, a diagnostic DT is responsible for troubleshooting and risk assessment.

\subsubsection*{Capability Level 3: Predictive}
Level 3 introduces predictive capabilities. Regarding ASVs, predicting the position and motion of other dynamic objects can be essential for proactive control. In addition, weather forecasts and predictive condition monitoring of vessel components can reduce the risk of critical outages during open-sea operations. While no predictive capabilities have been implemented yet, this work addresses this field by implementing a predictive target tracking (PTT) approach and a predictive safety filter (PSF), described in the following sections. The acquired knowledge from PTT extends the SITAW and can be used for the PSF for safe autonomous path following and COLAV.

\subsubsection*{Capability Level 4: Prescriptive}
The ability to predict the future states of the vessel, other dynamic objects, and environmental conditions allows the analysis of various what-if cases to provide an optimal decision. Previous work intimated this capability by applying deep reinforcement learning (DLR) for path following and collision avoidance, arguing that this learning method optimizes by exploring several generated scenarios. However, using purely machine learning-based methods might be critical for real-world applications since we cannot exactly interpret the generated output. Model predictive control (MPC) and its nonlinear equivalent NMPC are model-based predictive approaches optimizing the system behavior for a specific control horizon. The concept of PSF is based on NMPC and allows a comprehensive explanation of generated control inputs, which is later introduced and implemented in more detail, enabling first prescriptive capabilities.

\subsubsection*{Capability Level 5: Autonomous}
An autonomous DT is generally defined by a digital representative that closely maps the real system such that a loop from the DT to its real equivalent can be closed. As a result, the real system optimizes the DT, while the DT provides optimal decisions according to its enhanced SITAW.
For that purpose, adequate reliability of ASVs requires advanced and trustworthy control algorithms. The autonomous behavior of previous work was based on reinforcement learning, which led to potential collisions within hazardous scenarios containing highly dynamic objects.
In this work, we introduce the concept of a predictive safety filter (PSF) to guarantee safe control inputs generated by the DT.

\subsection{Vessel Model} \label{sec: Vessel_model}
The kinematical states of a surface vessel related to the global coordinates $[x, y]^\top$ are defined by $\boldsymbol{\eta} = [x_s,  y_s,  \psi]^\top$, where $x_s$ and $y_s$ are the $x$ and $y$ position of the ship, while $\psi$ denotes the heading.  Considering the velocity state vector $\boldsymbol{\nu} = [u,  v,  r]^\top$ containing surge $u$, sway $v$, and yaw rate $r$, the relation of $\boldsymbol{\nu}$ and $\boldsymbol{\eta}$ is given by
\begin{equation}
	\boldsymbol{\dot{\eta}}=\mathbf{R}_{\mathrm{rot}}(\psi)\boldsymbol{\nu},
\end{equation}
where $\mathbf{R}_{\mathrm{rot}}(\psi)$ denotes the rotational matrix, expressed by
\begin{equation}
	\mathbf{R}_{\mathrm{rot}}(\psi)=\begin{bmatrix}
		\mathrm{cos}(\psi) && -\mathrm{sin}(\psi) && 0\\
		\mathrm{sin}(\psi) && \mathrm{cos}(\psi) && 0 \\
		0 && 0 && 1
	\end{bmatrix}.
\end{equation}

According to \cite{Fossen2011homb}, the vessel dynamics can be described by 
\begin{equation}
	\mathbf{M}\boldsymbol{\dot{\nu}}+\mathbf{D}(\boldsymbol{\nu})\boldsymbol{\nu}+\mathbf{C}(\boldsymbol{\nu})\boldsymbol{\nu}=\boldsymbol{\tau}+\boldsymbol{\tau}_d, \label{eq: dynamics}
\end{equation}
Here, $\mathbf{M}$ specifies the mass matrix, $\mathbf{D}(\boldsymbol{\nu})$ is a nonlinear damping matrix, and $\mathbf{C} (\boldsymbol{\nu})$ is a Coriolis and centripetal matrix. Furthermore, $\boldsymbol{\tau}$ denotes the control input, while $\boldsymbol{\tau}_d$ characterizes the environmental disturbances impacted by wind, waves, and sea currents.
In control-theoretical notation, which is used later, the state space model yields
\begin{equation}
\hspace{-0.8em}\resizebox{.925\hsize}{0.022\vsize}{$\dot{\mathbf{x}} = \begin{bmatrix}
	    \mathbf{0}_{3x3} & \mathbf{R}_{\mathrm{rot}}(\mathbf{x})\\
            \mathbf{0}_{3x3} & \qquad  \mathbf{M}^{-1}[-\mathbf{C}(\mathbf{x})-\mathbf{D}(\mathbf{x})]
	\end{bmatrix}\mathbf{x}+\begin{bmatrix}
	    \mathbf{0}_{3x3}\\
            \mathbf{M}^{-1}
	\end{bmatrix} \mathbf{u}+\begin{bmatrix}
	    \mathbf{0}_{3x3}\\
            \mathbf{M}^{-1}
	\end{bmatrix} \mathbf{d}$}, \label{state_space_model}
\end{equation}
with the extended state vector $\mathbf{x}$, control input $\mathbf{u}$, and disturbances $\mathbf{d}$ given by
\begin{align}
    \mathbf{x}&=[x_s, y_s, \psi, u, v, r]^\top,\\
    \mathbf{u}&= \boldsymbol{\tau} = [\tau_u, \tau_v, \tau_r]^\top,\\
    \mathbf{d}&=\boldsymbol{\tau}_d=[\tau_{d,1},\tau_{d,2},\tau_{d,3}]^\top.   
\end{align}

\subsection{Nonlinear Model Predictive Control}
Model Predictive Control (MPC) is a proactive control method used in various complex fields, where it is important to handle constraints \cite{Biegler2021apo}. Its theory is extended to nonlinear model predictive control (NMPC), enabling one to deal with highly nonlinear systems.
The idea is to solve an optimization problem at each time step considering a set of hard and soft constraints. For that purpose, a cost function $J$ is minimized over a defined prediction horizon $N$.
A general discretized representation of the NMPC problem is given by
\begin{equation}
\hspace{-0.8em}
\begin{aligned}
& \underset{\mathbf{x}(t),\mathbf{u}(t)}{\text{min}}
& & J(\mathbf{x}(t), \mathbf{u}(t)) = \sum_{k=0}^{N-1} l(\mathbf{x}_k, \mathbf{u}_k) + V_f(\mathbf{x}_N) \\
& \ \ \ \text{s.t.}
& & \mathbf{x}_0 = \mathbf{x}(t), \\
&&& \mathbf{x}_{k+1} = f(\mathbf{x}_k, \mathbf{u}_k), \ \ \forall k \in [0, N-1] \\
&&& \mathbf{g}_h(\mathbf{x}_k, \mathbf{u}_k) \leq 0, \ \ \forall k \in [0, N-1]\\
&&& \mathbf{g}_s(\mathbf{x}_k, \mathbf{u}_k) \leq \mathbf{z}(k|t), \ \ \forall k \in [0, N-1]\\
&&& \mathbf{z}_k \geq 0, \ \ \forall k \in [0, N-1] \\
\end{aligned}
\end{equation}
The cost function consists of the stage cost $l(\mathbf{x}, \mathbf{u})$ and a terminal cost $V_f(\mathbf{x})$, where $k$ denotes the predicted timestep. The function $f(\mathbf{x}, \mathbf{u})$ express the system dynamics, and $\mathbf{g}_h(\mathbf{x}, \mathbf{u}) \leq 0$ characterize the hard constraints. Furthermore, $\mathbf{g}_s(\mathbf{x}, \mathbf{u}) \leq \mathbf{z}$ are the soft constraints, with slack variables $\mathbf{z}$ allowing for minor violations in constraints. In general, slack variables are additionally penalized in the cost function to minimize their usage. While hard constraints define strict boundaries on the system dynamics, soft constraints can relax the problem by enabling higher flexibility, which can be beneficial in the appearance of model uncertainties or if the optimal control problem (OCP) has no feasible solution. NMPC has proven to be capable of handling nonlinear systems despite its computational demands since permanent improvements in hardware have made NMPC applicable to various complex domains.

\subsection{Numerically Stable Ellipse Fitting} \label{sec: Ellipse_fitting_algorithm}
Algorithms for direct least squares fitting of ellipses using spatial data points are presented in \cite{Halir1998nsd, Fitzgibbon1999dls}, while \cite{Halir1998nsd} extends the method for improving numerical stability. In general, an ellipse can be defined by a second-order polynomial given by
\begin{equation}
F(x, y)=a x^2+b x y+c y^2+d x+e y+f=0, \label{eq:ellipse_equation}
\end{equation}
implying the constraint
\begin{equation}
b^2-4 a c<0. \label{eq:shape_fitting_unequality}
\end{equation}
The parameters $a$,$b$,$c$,$d$,$e$, and $f$ are coefficients of the ellipse, while $x$ and $y$ specify spatial coordinates.
Defining the two vectors
\begin{align}
& \mathbf{a}=[a, b, c, d, e, f]^\top, \label{eq:shape_fitting_coefficients}\\
& \boldsymbol{\xi}=\left[x^2, x y, y^2, x, y, 1\right],
\end{align}
leads to the expression
\begin{equation}
F_{\mathbf{a}}(\boldsymbol{\xi})=\boldsymbol{\xi} \cdot \mathbf{a}=0. \label{eq:ellipse_equation_compact}
\end{equation}
To find an optimal fitted ellipse to a set of data points, we try to find the argument $\mathbf{a}$ minimizing
\begin{equation}
\min _{\mathbf{a}} \sum_{i=1}^N F\left(x_i, y_i\right)^2 =\min _{\mathbf{a}} \sum_{i=1}^N\left(\boldsymbol{\xi}_{\mathbf{i}} \cdot \mathbf{a}\right)^2,
\end{equation}
resulting in a least square regression, where $N$ is the number of data points. According to \cite{Halir1998nsd}, there exists a freedom to arbitrarily scale the coefficients $\mathbf{a}$ such that instead of \eqref{eq:shape_fitting_unequality} we can apply the equality constraint  
\begin{equation}
4 a c-b^2 = 1.
\end{equation}
As a result, the optimization problem can be rewritten to
\begin{equation}
\begin{aligned}
&\min _{\mathbf{a}}\|\mathbf{D} \mathbf{a}\|^2,\\ 
& \text {s.t.} \quad \mathbf{a}^{\mathrm{T}} \mathbf{C} \mathbf{a}=1, \label{eq:shape_fitting_opt_prob}
\end{aligned}
\end{equation}
where the design matrix $\mathbf{D}$ is defined by
\begin{equation}
\mathbf{D}=\left[\begin{array}{cccccc}
x_1^2 & x_1 y_1 & y_1^2 & x_1 & y_1 & 1 \\
x_2^2 & x_2 y_2 & y_2^2 & x_2 & y_2 & 1 \\
\vdots & \vdots & \vdots & \vdots & \vdots & \vdots \\
x_n^2 & x_n y_n & y_n^2 & x_n & y_n & 1
\end{array}\right],
\end{equation}
and the constraint matrix $\mathbf{C}$ is given by
\begin{equation}
\mathbf{C}=\left[\begin{array}{cccccc}
0 & 0 & 2 & 0 & 0 & 0 \\
0 & -1 & 0 & 0 & 0 & 0 \\
2 & 0 & 0 & 0 & 0 & 0 \\
0 & 0 & 0 & 0 & 0 & 0 \\
0 & 0 & 0 & 0 & 0 & 0 \\
0 & 0 & 0 & 0 & 0 & 0
\end{array}\right].
\end{equation}
To solve the minimization problem, the quadratically constrained least squares minimization is proposed. Introducing the scatter matrix $\mathbf{S}=\mathbf{D}^\top \mathbf{D}$, and by applying the Lagrange multiplier, the conditions for an optimal solution of $\mathbf{a}$ yield
\begin{align}
    \mathbf{S}\mathbf{a}&=\lambda \mathbf{C} \mathbf{a}, \label{eq:shape_fitting_final_1}\\
    \mathbf{a}^\top \mathbf{C} \mathbf{a} &= 1. \label{eq:shape_fitting_final_2}
\end{align}
Generally, up to six solutions exist. However, considering \eqref{eq:shape_fitting_opt_prob}, \eqref{eq:shape_fitting_final_1}, and \eqref{eq:shape_fitting_final_2}, the identity
\begin{equation}
\|\mathbf{D} \mathbf{a}\|^2=\mathbf{a}^{\mathrm{T}} \mathbf{D}^{\mathrm{T}} \mathbf{D} \mathbf{a}=\mathbf{a}^{\mathrm{T}} \mathbf{S} \mathbf{a}=\lambda \mathbf{a}^{\mathrm{T}} \mathbf{C} \mathbf{a}=\lambda,
\end{equation}
is valid. This proves that the eigenvector $\mathbf{a}$ corresponding to the minimal positive eigenvalue $\lambda$ represents the best ellipse fit for a set of given data points.

\subsection{Kalman Filter}
Kalman filters (KFs) proved their capability in state estimation within several fields, including more than 20 modified filtering types, such as the extended Kalman filter (EKF) or the unscented Kalman filter (UKF) \cite{Chen2012kff}. 

Considering a discrete linear time-invariant (LTI) system defined by
\begin{align}
    &\mathbf{x}_{k+1} = \mathbf{A} \mathbf{x}_k + \mathbf{B} \mathbf{u}_k + \mathbf{w}_k, \\
    &\mathbf{z}_k = \mathbf{H} \mathbf{x}_k + \mathbf{v}_k,
\end{align}
where $\mathbf{A}$ defines the transition matrix, $\mathbf{B}$ is the control matrix, $\mathbf{H}$ characterizes the measurement matrix, $\mathbf{w}_k \sim \mathcal{N}(\mathbf{0},\mathbf{Q}_k)$ denotes the process noise with covariance matrix $\mathbf{Q}_k$, and $\mathbf{v}_k~\sim~\mathcal{N}(\mathbf{0},\mathbf{R}_k)$ is the measurement noise with covariance matrix $\mathbf{R}_k$.
The KF is a statistical filter using a prediction step and a correction step. The prediction step is defined by
\begin{align}
    &\check{\mathbf{x}}_{k|k-1} = \mathbf{A} \mathbf{x}_{k-1} + \mathbf{B} \mathbf{u}_{k-1}, \\
    &\check{\mathbf{P}}_{k|k-1} = \mathbf{A} \hat{\mathbf{P}}_{k-1} \mathbf{A}^\top + \mathbf{Q}_{k-1},
\end{align}
with the predicted (prior) state estimate $\check{\mathbf{x}}_{k|k-1}$, and the prior covariance matrix $\check{\mathbf{P}}_{k|k-1}$. The correction step of the KF is given by
\begin{align}
    &\tilde{\mathbf{y}}_k = \mathbf{z}_k - \mathbf{H} \check{\mathbf{x}}_{k|k-1}, \\
    &\mathbf{S}_k = \mathbf{H} \check{\mathbf{P}}_{k|k-1} \mathbf{H}^\top + \mathbf{R}_k, \\
    &\mathbf{K}_k = \check{\mathbf{P}}_{k|k-1} \mathbf{H} \mathbf{S}_k^{-1}, \\
    &\hat{\mathbf{x}}_{k|k} = \check{\mathbf{x}}_{k|k-1} + \mathbf{K}_k \tilde{\mathbf{y}}_k, \\
    &\hat{\mathbf{P}}_{k|k} = \left( \mathbf{I} - \mathbf{K}_k \mathbf{H} \right) \check{\mathbf{P}}_{k|k-1},
\end{align}
where $\tilde{\mathbf{y}}_k$ is the residual of the measurement and its model, $\mathbf{S}_k$ describes a cross-covariance matrix, $\mathbf{K}_k$ denotes the Kalman gain matrix, $\hat{\mathbf{x}}_{k|k}$ characterizes the updated (posterior) state estimate, and $\hat{\mathbf{P}}_{k|k}$ specifies the posterior covariance matrix. The more reliable the chosen model, the smaller $\check{\mathbf{P}}_{k|k-1}$, leading to a smaller Kalman gain $\mathbf{K}_k$, resulting in a higher weight of the model predictions $\check{\mathbf{x}}_{k|k-1}$.

\subsubsection*{Constant Velocity Model}
Using the nonlinear model dynamics of Section~\ref{sec: Vessel_model} to estimate and predict the state of other vessels and objects would require a nonlinear KF variant and the knowledge of unknown mass and hydrodynamic parameters. Since such information of other vessels is usually not accessible, we use a simplified constant velocity model for tracking other objects, commonly used in other studies \cite{Yue2023onp}. Considering the kinematics of a vessel and introducing the kinematical state vector $\mathbf{x}_{\mathrm{kin}} = \left[ x_s, y_s, \dot{x}_s, \dot{y}_s \right]^\top$, the constant velocity model can be defined by
\begin{equation}
    \mathbf{x}_{\mathrm{kin},k+1}=
    \begin{bmatrix}
        1 & 0 & \Delta t & 0 \\
        0 & 1 & 0 & \Delta t \\
        0 & 0 & 1 & 0 \\
        0 & 0 & 0 & 1
    \end{bmatrix} \mathbf{x}_{\mathrm{kin},k},
\end{equation}
where $\Delta t$ describes the time step between each update.

\subsection{Sensor Fusion}
\begin{figure}[htbp]
  \centering
  \includegraphics[width=1\linewidth]{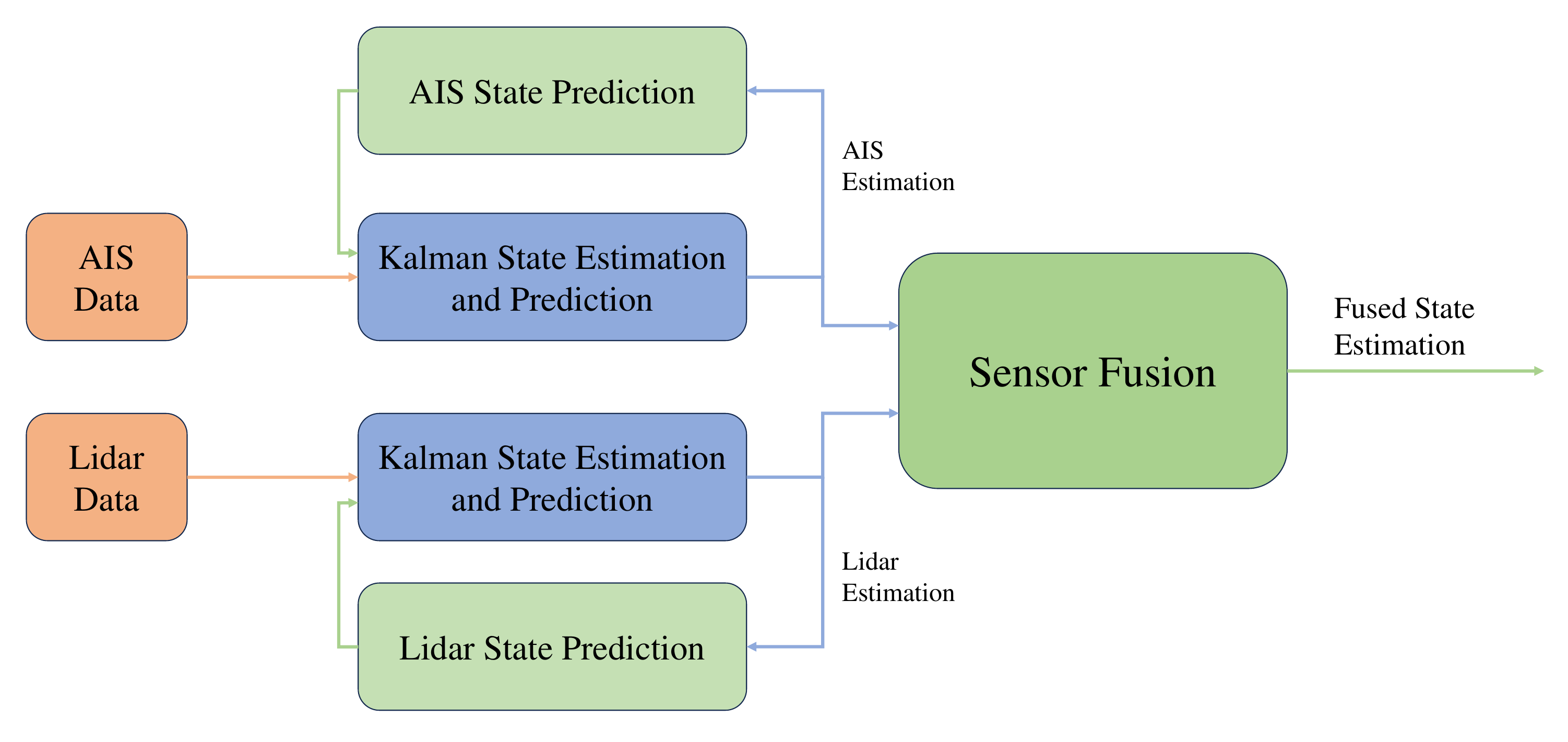} 
  \caption{Sensor fusion principle. AIS and LiDAR measurements are separately propagated through Kalman filters using a constant velocity model.}
  \label{fig:Sensor_fusion_parallel}
\end{figure}

Sensor fusion is a technique to combine the strengths of various sensors by simultaneously reducing their individual weaknesses. One of the most common approaches is Kalman filtering \cite{Fung2017sfa}.
In this work, the sensor fusion method used is inspired by \cite{Chen2019roa}, where AIS and radio detection and ranging (RADAR) data are fused using a late fusion principle. Late fusion is a technique that merges estimations and predictions of system states from multiple signal sources after applying their individual state estimations. This is usually done via Kalman filters and the signals are subsequently combined together via their related probability distributions to reduce the overall uncertainty.
An architecture based on a late fusion scheme is shown in Fig.~\ref{fig:Sensor_fusion_parallel}. 
According to \cite{Chen2019roa}, the prediction of the state resulting from the sensor fusion method is calculated by weighting the Kalman gain for the two models as follows
\begin{align}
&\mathbf{W}_{\mathrm{AIS}}=\mathbf{K}_{\mathrm{LiDAR},k+1} \left(\mathbf{K}_{\mathrm{AIS},k+1}+\mathbf{K}_{\mathrm{LiDAR},k+1}\right)^{-1}, \\
&\mathbf{W}_{\mathrm{LiDAR}}=\mathbf{K}_{\mathrm{AIS},k+1} \left(\mathbf{K}_{\mathrm{AIS},k+1}+\mathbf{K}_{\mathrm{LiDAR},k+1}\right)^{-1}, \\
&\mathbf{x}_{K+1}=\mathbf{W}_{\mathrm{AIS}} \mathbf{x}_{\mathrm{AIS},k+1}+\mathbf{W}_{\mathrm{LiDAR}} \mathbf{x}_{\mathrm{LiDAR},k+1}. \label{eq:sensor_fusion_literature}
\end{align}
Note that if the AIS Kalman gain is large, we assume that the AIS predictions from the constant velocity model are less trustworthy than the AIS measurements. Therefore, a large AIS Kalman gain leads to a stronger influence on the LiDAR predictions in \eqref{eq:sensor_fusion_literature} regarding the final state prediction. 

\section{Methodology} \label{sec:Method}
The novel contributions of this work build on an existing DT framework and extend it. Therefore, the required theory for the integrated target tracking approach and the concept of a predictive safety filter are theorized below. These approaches are later used to enhance the DT with respect to predictive, prescriptive, and improved autonomous capabilities.

\subsection{Predictive Target Tracking}
This section describes the implemented shape fitting approaches and the LiDAR-AIS sensor fusion concept used to estimate and predict the position of other objects.
\subsubsection*{LiDAR Shape Fitting}
Various LiDAR shape fitting approaches were compared in Python and Unity. However, within the Unity framework, only two offered a feasible solution. In addition to the method based on the ellipse fitting algorithm explained in Section \ref{sec: Ellipse_fitting_algorithm}, a shape fitting approach based on multiple linear regression (MLR) was implemented. MLR is applied to the ellipse equation \eqref{eq:ellipse_equation} to find the parameter vector $\mathbf{a}$ from \eqref{eq:ellipse_equation_compact}.
The objective of MLR is to estimate the coefficients that minimize the sum of squared differences between the observed and predicted values. However, the algorithm finally used in the DT is the numerically stable ellipse fitting algorithm described in the theory section due to accuracy and computational cost benefits.

\subsubsection*{LiDAR-AIS Sensor Fusion}
The fusion model \eqref{eq:sensor_fusion_literature} presented in \cite{Chen2019roa} uses Kalman gains to weight the individual contributions. In this study, the individual measurements $\mathbf{z}_i$ and covariance matrices $\mathbf{R}_i$ are used to fuse the sensors and reduce the overall uncertainty. 
The AIS and LiDAR measurements are fused after performing a Kalman filter using a constant velocity model.
Since AIS data are received approximately once per minute, while LiDAR measurements arrive with a high frequency, sensor fusion only applies if AIS data are registered, allowing for a simple synchronization.
Considering the individual measurement sources $\mathbf{z}_i$ and their dedicated covariance matrices $\mathbf{R}_i$, the fusion model is defined by a probability distribution formulated by

\begin{equation}
\hspace{-0.9em}\resizebox{.93\hsize}{0.04\vsize}{$
\mathcal{N}(\hat{\mathbf{x}},\hat{\mathbf{P}}) = \begin{cases}
\mathcal{N}(\check{\mathbf{x}},\check{\mathbf{P}})\mathcal{N}(\mathbf{z}_{\mathrm{LiDAR}}, \mathbf{R}_{\mathrm{LiDAR}}), & \text{if } \mathbf{z}_{\mathrm{LiDAR}} \land \neg \mathbf{z}_{\mathrm{AIS}} \\
\mathcal{N}(\check{\mathbf{x}},\check{\mathbf{P}})\mathcal{N}(\mathbf{z}_{\mathrm{AIS}}, \mathbf{R}_{\mathrm{AIS}}), & \text{if } \neg \mathbf{z}_{\mathrm{LiDAR}} \land \mathbf{z}_{\mathrm{AIS}} \\
\mathcal{N}(\check{\mathbf{x}},\check{\mathbf{P}})\mathcal{N}(\mathbf{z}_{\mathrm{LiDAR}}, \mathbf{R}_{\mathrm{LiDAR}}) \mathcal{N}(\mathbf{z}_{\mathrm{AIS}}, \mathbf{R}_{\mathrm{AIS}}), & \text{if } \mathbf{z}_{\mathrm{LiDAR}} \land \mathbf{z}_{\mathrm{AIS}} \\
\mathcal{N}(\check{\mathbf{x}},\check{\mathbf{P}}), & \text{if } \neg \mathbf{z}_{\mathrm{LiDAR}} \land \neg \mathbf{z}_{\mathrm{AIS}}
\end{cases}$} \label{eq:Sensor_fusion}
\end{equation}
Note that the logical expressions related to the measurement sources $\mathbf{z}_i$ specify whether measurements are received.

\subsection{Predictive Safety Filter}
\begin{figure}[htb]
    \centering  \includegraphics[width=\linewidth]{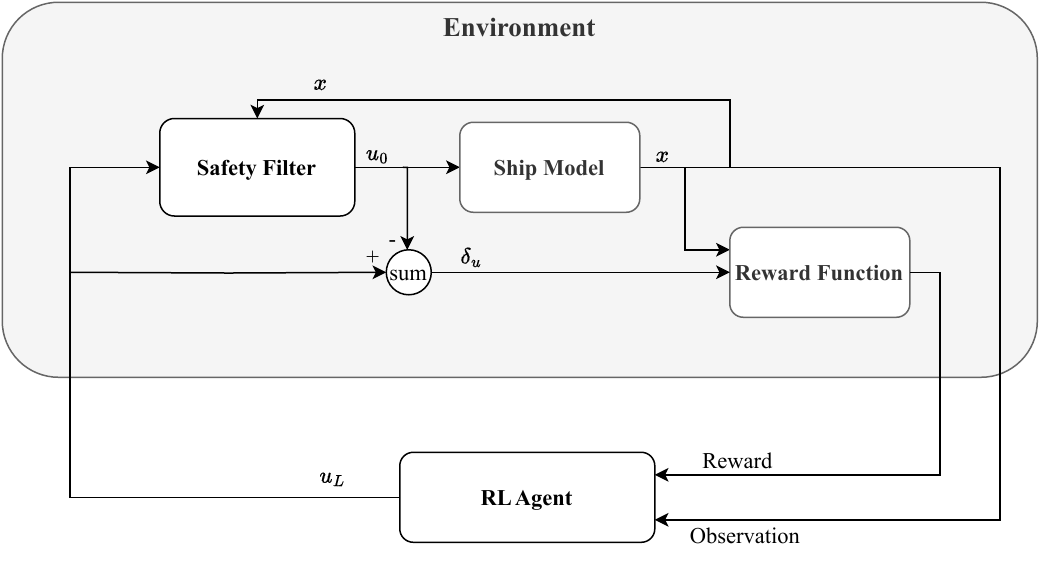}
    \caption{Illustration of the RL and PSF control design \cite{Vaaler2023mca}.}
    \label{fig:RL_PSF_fig}
\end{figure}

The concept of a predictive safety filter (PSF) for learning-based control of constrained nonlinear dynamical systems was first introduced by \cite{Wabersich2021aps} and is inspired by the theory of NMPC. A PSF receives control inputs from a learning-based controller, often obtained by reinforcement learning (RL), and verifies if the control input can be safely executed. Otherwise, the control input is modified. In \cite{Vaaler2023mca}, a PSF coupled with RL is proposed for safe maritime navigation of ASVs, showing promising results in terms of navigation and control. Therefore, this methodology is used to extend the DT by generating safe control inputs and interfacing them with the Unity framework, where the predicted safety path is visualized. The control design is demonstrated in Fig.~\ref{fig:RL_PSF_fig}, where the objective is to minimize the difference between the control input generated by the RL agent $\mathbf{u}_L$ and a safe control input $\mathbf{u}_0$ obtained from the PSF. 
If the safety filter needs to interfere, the RL agent adopts according to an adapted reward function manipulated by 
\begin{equation}
   \boldsymbol{\delta}_u = \mathbf{u}_L - \mathbf{u}_0. 
\end{equation}
 
Therefore, the PSF tries to find a minimal perturbed control action that guarantees safety over a specific prediction horizon $N$. For that purpose, a safety distance $d_{safe}$ concerning the distance $d(\mathbf{p},\mathbb{O}_i)$ from the vessel's position $\mathbf{p} = [x_s, y_s]^\top$ and a set of obstacle positions $\mathbb{O}_i$ must be guaranteed. As a result, the OCP is formulated by

\begin{equation}
\begin{aligned}
& \underset{\mathbf{u}(t), \mathbf{x}(t)}{\text{min}}
& & \sum_{k=0}^{N-1} ||\mathbf{u}_{0,k}-\mathbf{u}_{L,k}||^2_{\mathbf{W}}   \\ \\
& \ \ \ \text{s.t.}
& &  (a) \quad \mathbf{x}_{k+1} = f(\mathbf{x}_k, \mathbf{u}_k, \mathbf{d}_k), \ \ \forall k \in [0, N - 1]\\
& & & (b) \quad \mathbf{x}_0 = \mathbf{x}(t) \\
& & & (c) \quad \mathbf{x}_{lb} \leq \mathbf{x}_k \leq \mathbf{x}_{ub},  \ \ \forall k \in [0, N] \\
& & & (d) \quad \mathbf{u}_{lb} \leq \mathbf{u}_k \leq \mathbf{u}_{ub},  \ \ \forall k \in [0, N - 1]\\
& & & (e) \quad d(\mathbf{p}_{k},\mathbb{O}_{i,k}) \geq d_{safe},  \ \ \forall k \in [0, N - 1] \\
& & & (f) \quad d(\mathbf{p}_{N},\mathbb{O}_{i,N}) \geq d_{safe} + d_{f}, \\
& & & (g) \quad \boldsymbol{\nu}_{N}^T\mathbf{P}_{f_{\boldsymbol{\nu}}}\boldsymbol{\nu}_{N} \leq 1 
\end{aligned} \label{eq:OCP}
\end{equation}

\begin{figure}[htbp]
    \centering    \includegraphics[width=\linewidth]{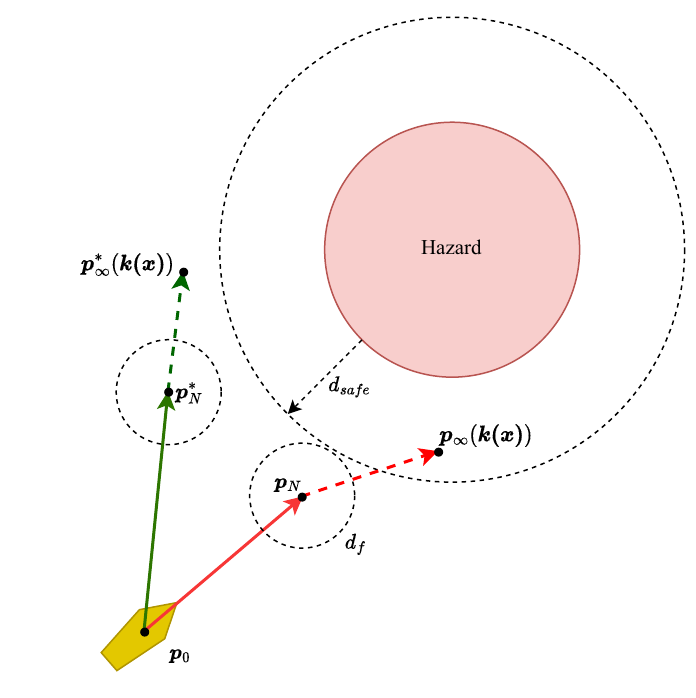}
    \caption[Terminal safety constraint visualization]{Visualization of ship trajectory modification caused by terminal safety constraint, with $N=1$ for clarity. The red arrows indicate the nominal (unsafe) trajectory, while the green arrows indicate the modified PSF trajectory \cite{Vaaler2023mca}.}
    \label{fig:ship_PSF_visual}
\end{figure}

Note that $(a)$ is given by the state space model \eqref{state_space_model}, $\mathbf{u}_{0}$ is the control input provided by the PSF, and $\mathbf{u}_{L}$ is the learning-based control input. The lower and upper bounds of the states and control inputs are denoted as $\mathbf{x}_{lb}$, $\mathbf{x}_{ub}$, $\mathbf{u}_{lb}$, $\mathbf{u}_{ub}$, respectively.
Let $d_f$ be a feasible safety distance of the terminal position $\mathbf{p}_N$, we additionally ensure safety with $(f)$ beyond the prediction horizon $N$. Figure~\ref{fig:ship_PSF_visual} demonstrates the concept of introducing a terminal safe set $\mathbb{X}_f$ to guarantee $\mathbf{x}_\infty \in \mathbb{X}_f$ with a feasible safety distance $d_f$. 
The constraint $(g)$ ensures that $\mathbf{x}$ remains within a control invariant ellipsoidal set $\mathbb{X}_e$ for $t \in (t_N,\infty)$ with $\mathbb{X}_e \in \mathbb{X}_f$ and $\mathbb{X}_e := \{ \mathbf{x} | \mathbf{x}^\top \mathbf{P}\mathbf{x} \leq 1 \}$. Therefore, $\mathbf{P}$ must be positive definite. In \eqref{eq:OCP}, $\mathbf{P}_{f_{\boldsymbol{\nu}}}$ is related to the velocity vector $\boldsymbol{\nu}$ given by
\begin{equation}
    \mathbf{P}_{f_{\boldsymbol{\nu}}} =\begin{bmatrix}       p_{44} & p_{45} & p_{46} \\
        p_{54} & p_{55} & p_{56} \\
        p_{64} & p_{65} & p_{66}\end{bmatrix}.
\end{equation}
The matrix $\mathbf{P}_{f_{\boldsymbol{\nu}}}$ can be computed by constructing a semidefinite program (SDP) which is described in more detail in \cite{Wabersich2021aps, Vaaler2023mca, Verschueren2022aam}. In addition, the matrix $\mathbf{W}$ weighting the cost function is defined by
\begin{equation}
    \mathbf{W} = \begin{bmatrix}
        \frac{\gamma_{u_1}}{(u_{1,ub}-u_{1,lb})^2} & 0 & 0\\
        0 & \frac{\gamma_{u_2}}{(u_{2,ub}-u_{2,lb})^2} & 0\\
        0 & 0 & \frac{\gamma_{u_3}}{(u_{3,ub}-u_{3,lb})^2}\end{bmatrix},
\end{equation}
leading to a normalization regarding the control input boundaries, where $\gamma_{u_1}$, $\gamma_{u_2}$, and $\gamma_{u_3}$ are tuning parameters.

\subsubsection*{Reward Function}
The reward function used for the RL agent consists of three components. A path following reward, a COLAV reward, and a safety violation reward. The path following and the COLAV reward are chosen from the proposal of \cite{Meyer2020ccc, Meyer2020taa}. In this regard, path following is rewarded by
\begin{equation}
r_{\text {path }}=\underbrace{\left(\frac{u}{U_{\max}} \cos \bar{\psi}+\gamma_r\right)}_{\text {Velocity-based reward }} \underbrace{\left(\exp \left(-\gamma_\epsilon\left|\epsilon\right|\right)+\gamma_r\right)}_{\text {CTE-based reward }}-\gamma_r^2,
\end{equation}
where the velocity-based reward contains the current surge speed $u(t)$, the maximum speed $U_{\max}$, the heading error $\bar{\psi}(t)$, and a tuning parameter $\gamma_r$ to avoid a small reward if the cross-track error (CTE) is large. The CTE-based reward implies the CTE $\epsilon(t)$, and a further tuning parameter $\gamma_\epsilon$. COLAV is rewarded by the weighted average of the LiDAR measurements given by

\begin{equation}
r_{\text {colav }}=-\frac{\sum_{i=1}^n \frac{1}{1+\gamma_\theta\left|\theta_i\right|} \exp \left(-\gamma_d d_i\right)}{\sum_{i=1}^n \frac{1}{1+\gamma_\theta\left|\theta_i\right|}},
\end{equation}
where $n$ is the number of LiDAR measurements, $d_i$ are the dedicated distances, $\theta_i$ characterize the relative angle of the tracked vessel, while $\gamma_\theta$ and $\gamma_d$ are tuning parameters.

The safety violation reward is defined by
\begin{equation}
    r_{\mathrm{PSF}} =-\gamma_{\mathrm{PSF}} \frac{||\mathbf{u}_{\mathrm{L}}-\mathbf{u}_0||}{||\mathbf{u}_{\max }||} , \label{eq:PSF_reward}
\end{equation}
with tuning parameter $\gamma_{\mathrm{PSF}}$. The maximum control input $\mathbf{u}_{\max}$ is used in \eqref{eq:PSF_reward} for normalization.
As a result, the entire reward is formulated by
\begin{equation}
r= \begin{cases}r_{\text {collision }}, & \text { if collision } \\
\lambda r_{\text {path }}+(1 - \lambda) r_{\text {colav }}+ r_{\mathrm{PSF}}+ r_{\text {exists }}, & \text { otherwise }\end{cases}
\end{equation}
where $r_{\text {collision}}$ defines a large negative reward in case of a collision, $r_{\text{exist}}$ is a constant negative reward, and $\lambda$ denotes a tuning parameter weighting the tradeoff regarding path following and COLAV.

\section{Results and Discussion} \label{sec:Results}
In general, the DT built with the Unity game engine demonstrates the ability to model, predict, and make optimal decisions in a safe virtual environment. Using real-world data, such as AIS data, extends the SITAW and allows model corrections and predictions by applying advanced tracking algorithms.

\begin{figure}[htpb]
  \centering
  \begin{subfigure}{0.495\linewidth}
    \centering
    \adjustbox{trim=0cm 0cm 0.4cm 0cm, clip=true}{\includegraphics[width=\linewidth]{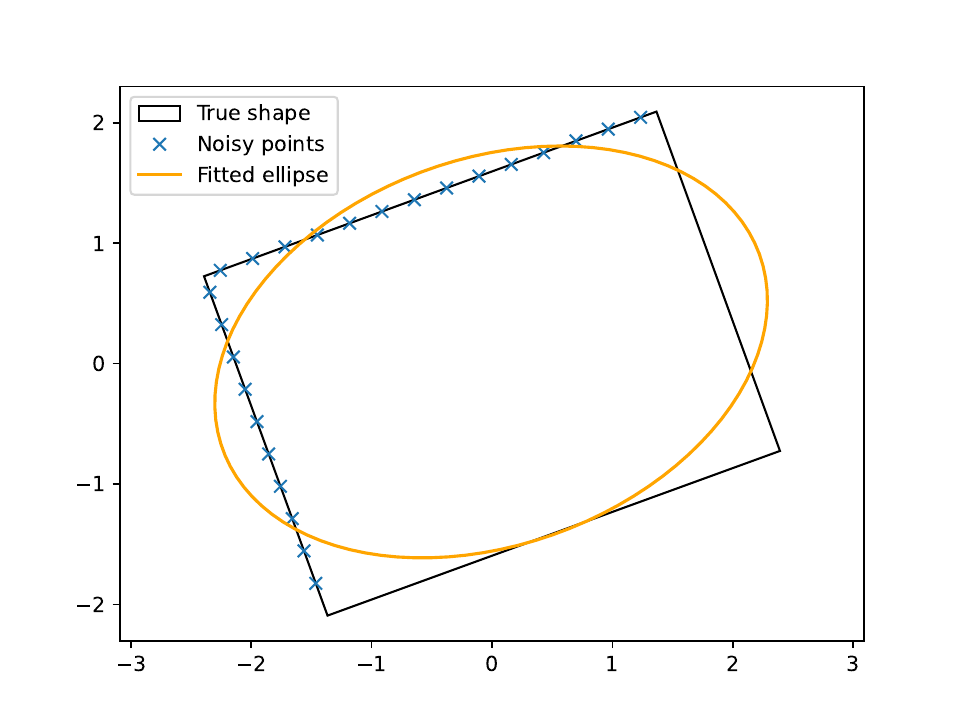}}
    \caption{Without noise}
    \label{StabelRectFittingWithE}
  \end{subfigure}
  \begin{subfigure}{0.495\linewidth}
    \centering
    \adjustbox{trim=0cm 0cm 0.4cm 0cm, clip=true}{\includegraphics[width=\linewidth]{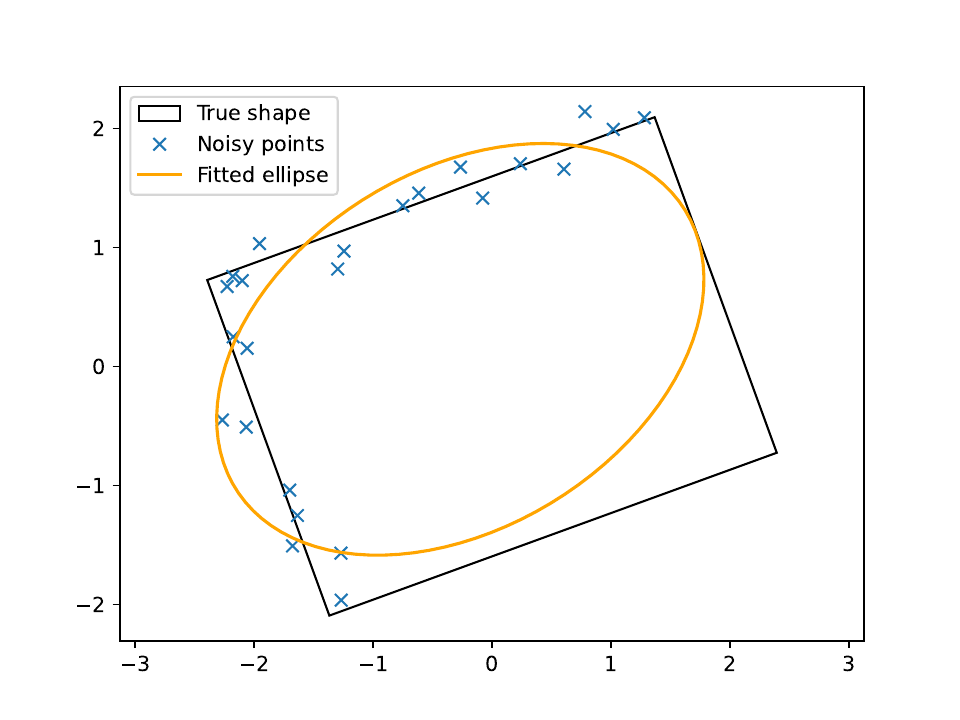}}
    \caption{With noise}
    \label{StabelRectFittingWithEN}
  \end{subfigure}
  \caption{Fitting an ellipse to a rectangle with the numerically stable direct least squares approach. The blue points demonstrate the LiDAR measurements. As presented in Fig~\ref{StabelRectFittingWithEN}, the algorithm is very robust even when noise is added to the measurements.}
  \label{fig:RectEllipseS}
\end{figure}

\begin{figure}[htbp]
	\centering
	\includegraphics[width=\linewidth]{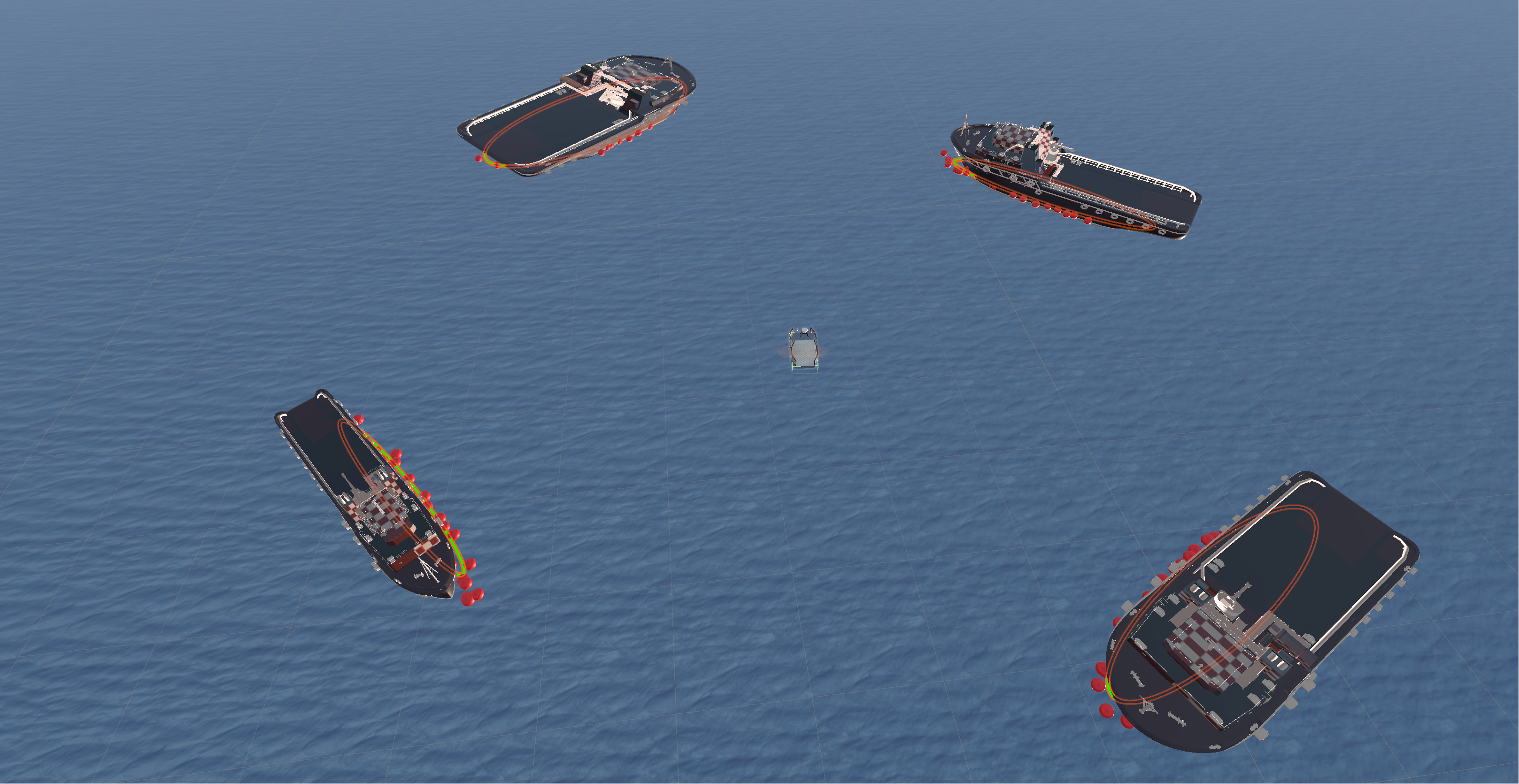}
	\caption{Fitting an ellipse within the DT framework built in Unity using the numerically stable ellipse fitting algorithm. The synthetical LiDAR measurements, depicted as red dots, are simulated with noise. In this visualization, the fitted ellipses are shown in orange and demonstrate the capability of tracking multiple targets in parallel.}
	\label{fig:ellipse_fitting_in_Unity}
\end{figure}

The numerically stable direct least squares ellipse fitting algorithm is depicted in Fig.~\ref{fig:RectEllipseS}.
Compared are the results considering perfect measurements and noisy measurements. 
It can be seen that, even if the measurements are unreliable, the algorithm can provide a sufficient shape estimation.
Furthermore, it is demonstrated that the algorithm can fit an ellipse even if the shape is edged like a rectangular, which appears on some surface vessels.
The implementation in the DT framework is depicted in Fig.~\ref{fig:ellipse_fitting_in_Unity}. Presented is a target tracking scenario of multiple objects (vessels) where the synthetically generated LiDAR point clouds contain noise.
The noisy measurements are depicted as red dots, while fitted ellipses are colored orange.
The fitted shapes additionally enable pose estimations due to the orientation and motion of the ellipses, which can be useful for predicting the paths of other vessels. Especially in multi-target tracking scenarios using LiDAR point clouds, such capabilities can guarantee improved predictive reliabilities if multiple objects are merged within a single point cloud.


\begin{figure}[htbp!]
	\centering
	\includegraphics[width=\linewidth]{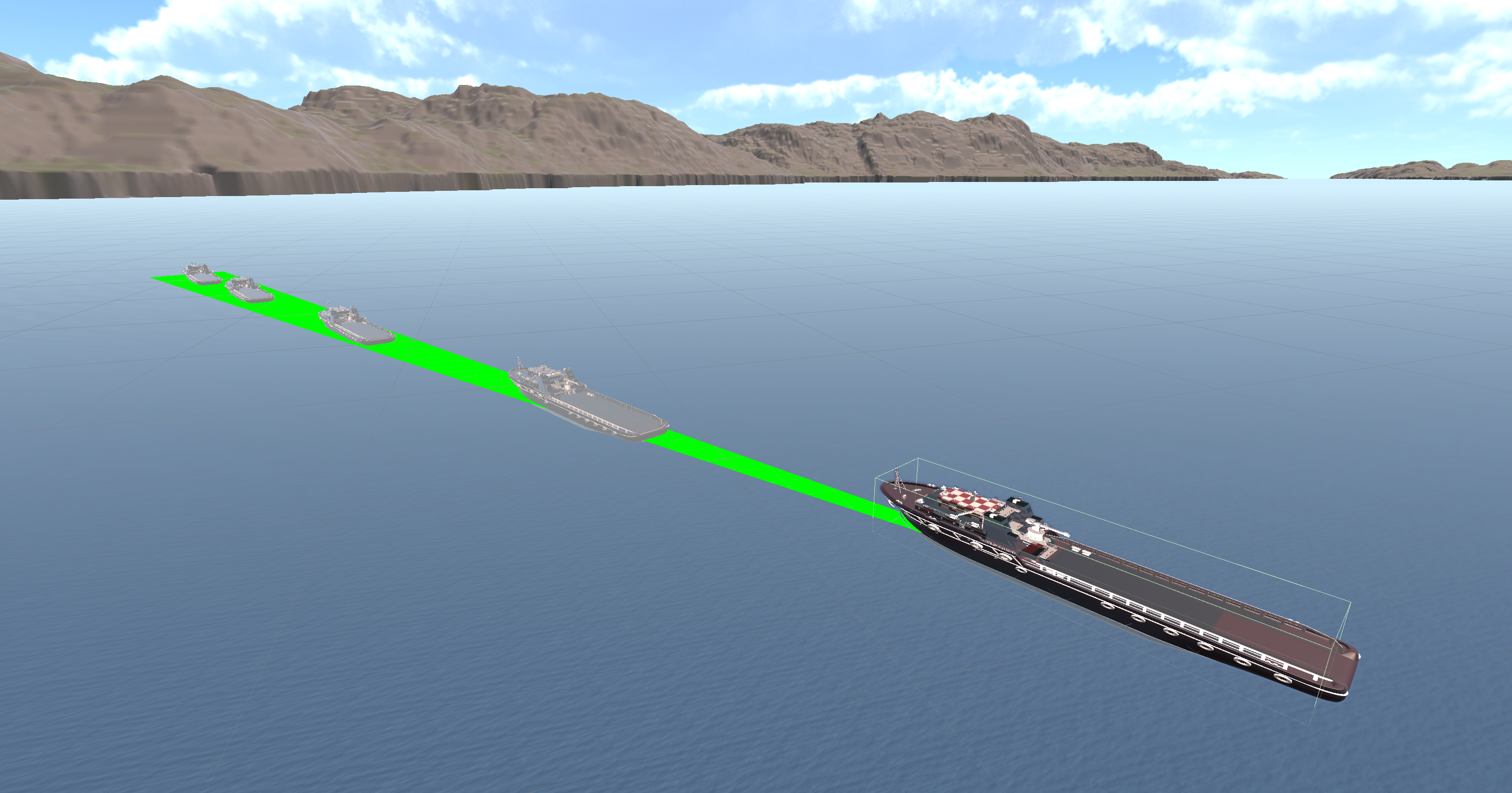}
	\caption{Prediction of the position and motion of a tracked vessel within the DT. The green broadening path demonstrates the increasing uncertainty of the prediction.}
	\label{fig:DT_Kalman_predictions}
\end{figure}

The predictions of the sensor fusion method are demonstrated in Fig.~\ref{fig:DT_Kalman_predictions}, where the green broadening path represents the predicted path with cumulative uncertainty. 
\begin{figure*}[htbp!]
\centering
\begin{subfigure}[h]{0.32\linewidth}
\centering
\includegraphics[width=\linewidth]{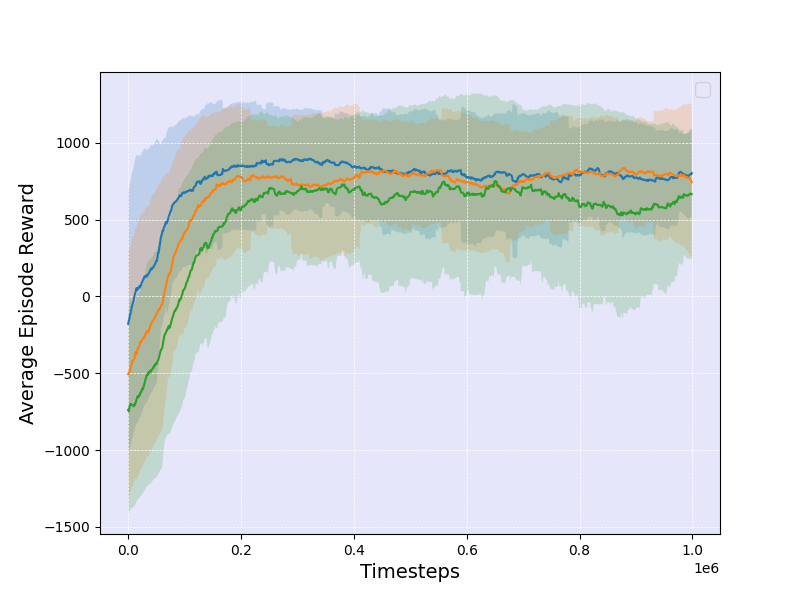}
\subcaption{Reward}
\label{fig:case2_reward}
\end{subfigure}
\begin{subfigure}[h]{0.32\linewidth}
\centering
\includegraphics[width=\linewidth]{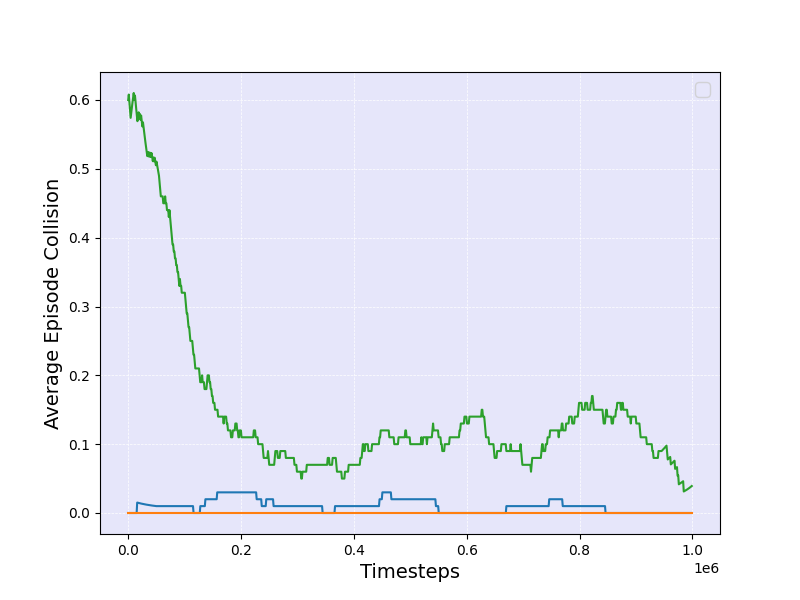}
\subcaption{Collision rate}
\label{fig:case2_collision}
\end{subfigure}
\begin{subfigure}[h]{0.32\linewidth}
\centering
\includegraphics[width=\linewidth]{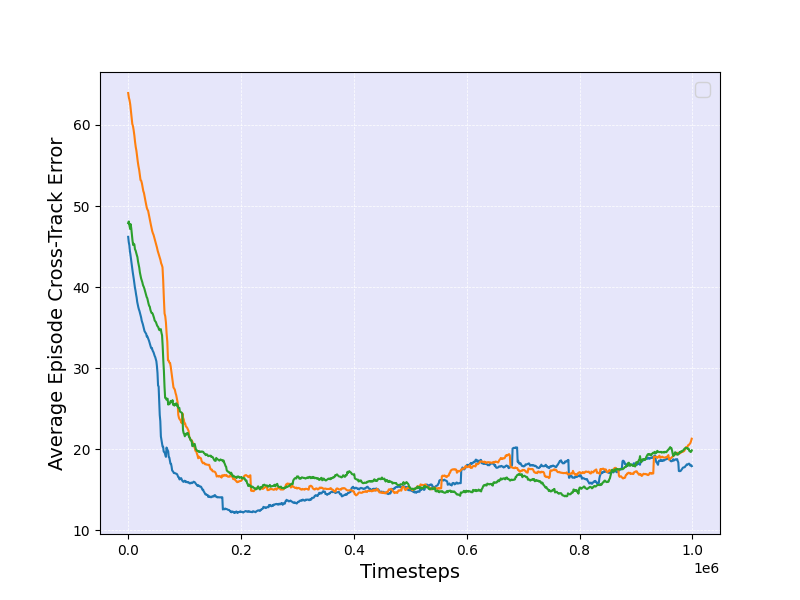}
\subcaption{Cross-track error}
\label{fig:case2_cte}
\end{subfigure}
\begin{subfigure}[h]{\linewidth}
\centering
\includegraphics[width=0.5\linewidth]{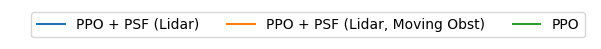}
\end{subfigure}
\caption{Average reward, collisions, and cross-track error during training smoothed with a rolling average over 100 episodes. Only the PPO+PSF with information about the moving obstacles has a collision rate of zero during the entire training period.}
\label{fig:PSF_PPO_training}
\end{figure*}
During open-sea operations, the tracking predictions of the DT were relatively accurate. However, during close-to-coast operations, it turned out that the predictions can turn unstable. This issue is most likely related to the integration within the Unity game engine. Reasons could be restrictions related to the used CV model or Unity-specific limitations such as Unity's graphics and physics integration related to rendering, or the Unity application programming interface (API).

Extended to the work presented in \cite{Menges2023dtf}, the RL-driven ASV in the DT is equipped with a PSF, allowing safe path predictions. In Fig.~\ref{fig:PSF_PPO_training}, the training process of the RL agent is depicted. For training, proximal policy optimization (PPO) was applied. Therefore, we compared three different setups. PPO without PSF, PPO with PSF, and PPO with PSF, where the PSF was instilled with additional information about the motion of the obstacles.
Presented are the average reward per episode, the average collision rate, and the average CTE with a rolling average over 100 episodes. It can be seen that the additional use of a PSF improves the COLAV rate drastically, which is the main goal for safe maritime navigation. A more detailed analysis of the approach is given in \cite{Vaaler2023mca}.

\begin{figure}[htbp]
	\centering
	\includegraphics[width=\linewidth]{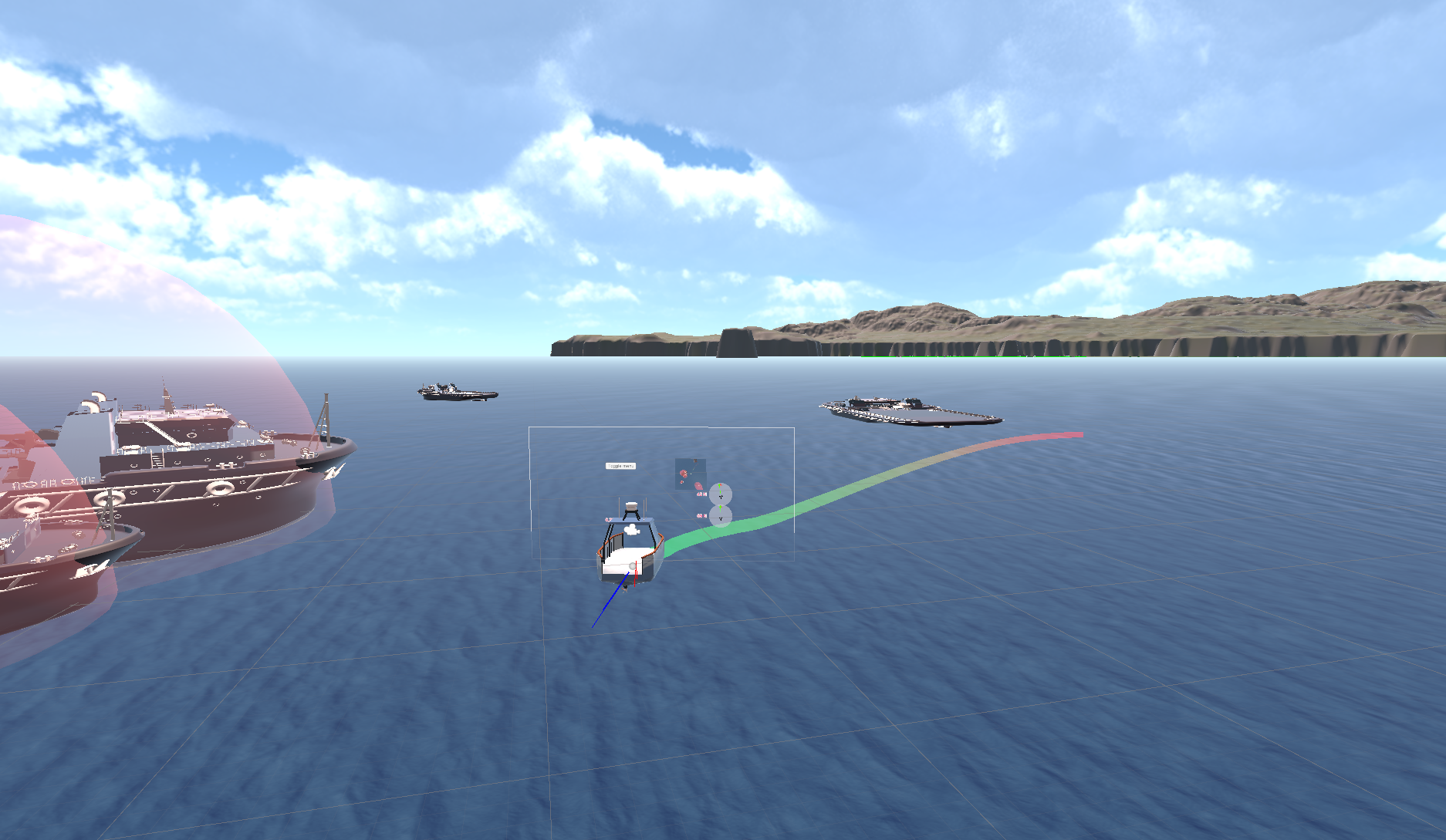}
	\caption{Demonstration of the PSF in the DT framework. The green path visualizes the predicted safe path, with critical regions depicted in red.}
	\label{fig:safety_filter_DT}
\end{figure}

The implementation of the PSF within the DT is demonstrated in Fig.~\ref{fig:safety_filter_DT}. 
The predicted safe path is shown in green, while the red ending shows a critical predicted unsafe region. Such predictive capabilities can reduce risk and guarantee the safety of potential real-world applications. Since this DT is not connected to the real-world ASV, no loop is closed yet. However, the work shows, in general, what a DT for ASVs and their environment could be capable of. As shown in the previous work, other real-time sensor data, such as AIS and weather data, are streamed into the DT. These capabilities facilitate an extended SITAW for improved risk assessment and safer control.

\section{Conclusion and Future Work} \label{sec:Conclusion}
Considering the capability scale of a DT presented in Section~\ref{sec:Digital_twins}, the existing DT framework is extended by the first predictive and prescriptive capabilities. In addition to the numerically stable ellipse fitting approach for other objects, predictive target tracking using Kalman filters and the probabilistic sensor fusion technique given in \eqref{eq:Sensor_fusion} allows for path predictions of other vessels, including their cumulative uncertainty estimate. Furthermore, the predictive safety filter (PSF) integrates an additional security factor into the DT, enabling proactive SITAW and COLAV of ASVs. These integrations can guarantee safer sea operations through the prescriptive analysis of what-if scenarios, allowing the RL-driven control approach for a significant risk reduction. Even if several capabilities are already integrated into the DT,  each individual capability level has the potential for improvements and extensions. Additional physics and data-driven models, additive data sources for model corrections, and more intelligent algorithms are planned to be integrated in the future.
\\
	
	
\section*{Acknowledgments}
This work is part of SFI AutoShip, an 8-year research-based innovation center. 
In addition, this research project is integrated into the PERSEUS doctoral program. 
We want to thank our partners, including the Research Council of Norway, under project number 309230, and the European Union’s Horizon 2020 research and innovation program under the Marie Skłodowska-Curie grant agreement number 101034240.

\bibliographystyle{AR}
\bibliography{references}
	
	
\appendix

	

\end{document}